\documentclass[11pt,a4paper]{article}
\usepackage[hyperref]{emnlp-ijcnlp-2019}
\usepackage{times}
\usepackage{latexsym}

\usepackage{url}
\usepackage{amsmath}
\usepackage{algorithm}
\usepackage{algpseudocode}
\usepackage{graphicx}
\usepackage{tabularx}
\aclfinalcopy 

\newcolumntype{L}[1]{>{\raggedright\arraybackslash}p{#1}}
\newcolumntype{C}[1]{>{\centering\arraybackslash}p{#1}}
\title{A Benchmark Dataset for Learning to Intervene in Online Hate Speech}
\author{Jing Qian$^\dagger$, 
Anna Bethke$^\ast$, 
Yinyin Liu$^\ast$, 
Elizabeth Belding$^\dagger$, 
William Yang Wang$^\dagger$\\
$^\dagger$ University of California, Santa Barbara\\
$^\ast$ Intel AI \\
  {\tt \{jing\_qian,ebelding,william\}@cs.ucsb.edu} \\ \tt \{anna.bethke,yinyin.liu\}@intel.com  \\}

\date{}

\begin{document}
\maketitle
\begin{abstract}
  
\end{abstract}
Countering online hate speech is a critical yet challenging task, but one which can be aided by the use of Natural Language Processing (NLP) techniques. Previous research has primarily focused on the development of NLP methods to automatically and effectively detect online hate speech while disregarding further action needed to calm and discourage individuals from using hate speech in the future. In addition, most existing hate speech datasets treat each post as an isolated instance, ignoring the conversational context. In this paper, we propose a novel task of generative hate speech intervention, where the goal is to automatically generate responses to intervene during online conversations that contain hate speech. As a part of this work, we introduce two fully-labeled large-scale hate speech intervention datasets\footnote{https://github.com/jing-qian/A-Benchmark-Dataset-for-Learning-to-Intervene-in-Online-Hate-Speech} collected from Gab\footnote{https://gab.ai} and Reddit\footnote{https://www.reddit.com}. These datasets provide conversation segments, hate speech labels, as well as intervention responses written by  Mechanical Turk\footnote{https://www.mturk.com} Workers. In this paper, we also analyze the datasets to understand the common intervention strategies and explore the performance of common automatic response generation methods on these new datasets to provide a benchmark for future research. 

\begin{figure}[t]
    \centering
    \includegraphics[width=1.0\linewidth]{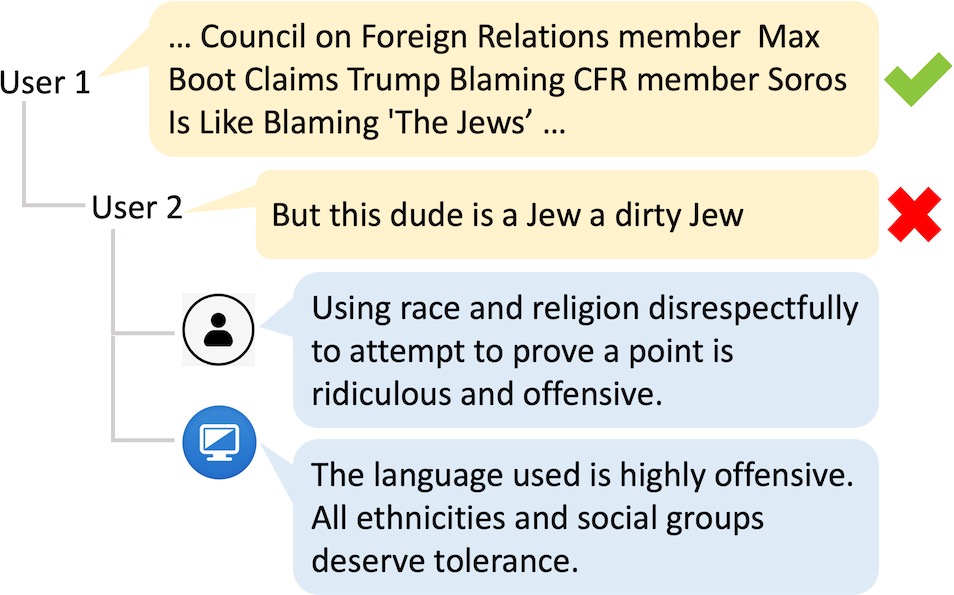}
    \caption{An illustration of hate speech conversation between User 1 and User 2 and the interventions collected for our datasets. The check and the cross icons on the right indicate a normal post and a hateful post. The utterance following the human icon is a human-written intervention, while the utterance following the computer icon is  machine-generated.}
    \label{fig:intro}
\end{figure}

\section{Introduction}
\label{sec:introduction}
The growing popularity of online interactions through social media has been shown to have  both positive and negative impacts. While social media improves information sharing, it also  facilitates the propagation of online harassment, including hate speech. These negative experiences can have a measurable negative impact on users. Recently, the Pew Research Center~\cite{pew2017online} reported that ``roughly four-in-ten Americans have personally experienced online harassment, and 63\% consider it a major problem.''

To address the growing problem of  online hate, an extensive body of work has  focused on developing automatic hate speech detection models and datasets~\cite{warner2012detecting,waseem2016hateful,davidson2017automated, shmidt2017survey, elsherief2018hate,elsherief2018peer,qian2018hierarchical,qian2018leveraging}. However, simply detecting and blocking hate speech or suspicious users often has limited ability to prevent these users from simply turning to other social media platforms to continue to engage in hate speech as can be seen in the large move of individuals blocked from Twitter\footnote{https://www.twitter.com} to Gab~\cite{ohlheiser2016}. What's more, such a strategy is often at odds with the concept of free speech. As reported by the Pew Research Center~\cite{pew2017online}, ``Despite this broad concern over online harassment, 45\% of Americans say it is more important to let people speak their minds freely online; a slightly larger share (53\%) feels that it is more important for people to feel welcome and safe online.'' 
The special rapporteurs representing the Office of the United Nations High Commissioner for Human Rights (OHCHR) have recommended that ``The strategic response to hate speech is more speech.''~\cite{ohchr2011} They encourage to change what people think instead of merely changing what they do, so they advocate more speech that educates about cultural differences, diversity, and minorities as a better strategy to counter hate speech.

Therefore, in order to encourage strategies of countering online hate speech, we propose a novel task of generative hate speech intervention and introduce two new datasets for this task. Figure~\ref{fig:intro} illustrates the task. Our datasets consist of 
5K conversations retrieved from Reddit and 12k conversations retrieved from Gab. 
Distinct from existing hate speech datasets, our datasets retain their conversational context and introduce human-written intervention responses. The conversational context and intervention responses are critical in order to build generative models to automatically mitigate the spread of these types of conversations. 

To summarize, our contributions are three-fold:
\begin{itemize}
    \item We introduce the generative hate speech intervention task and provide two fully-labeled hate speech datasets with human-written intervention responses.
    \item Our data is collected in the form of conversations, providing better context.
    \item The two data sources, Gab and Reddit, are not well studied for hate speech. Our datasets fill this gap.
\end{itemize}

 Due to our data collecting strategy, all the posts in our datasets are manually labeled as 
 hate or non-hate speech by Mechanical Turk workers, so they can also be used for the hate speech detection task. The performance of commonly-used classifiers on our datasets is shown in Section~\ref{sec:experimens}.

\begin{table*}[t!]
\centering
\small
\begin{tabular}{|l|C{20mm}|c|c|p{48mm}|c|}
  \hline
     & {Source} &\#Posts  & Conv. &\multicolumn{1}{|c|}{Categories} & Interv. \\
  \hline
  \citet{waseem2016hateful} &{Twitter} &17k  &No &racist, sexist, normal  &No  \\
  \hline
  \citet{davidson2017automated}&{Twitter} &25k  &No &hateful, offensive, neither &No \\
  \hline
  \citet{golbeck2017large}&{Twitter} &35k  &No &the worst, threats, hate speech, direct harassment, potentially offensive, non-harassment &No\\
  \hline
  \citet{chatzakou2017mean} &{Twitter} &9k  &No &aggressive, bullying, spam, normal &No\\
  \hline
  \citet{george2017technology}  &{Twitter, Reddit, The Gaurdian} &20k  &No &harassment, non-harassment &No \\
  \hline
  \citet{founta2018large}  &Twitter &100k  &No &abusive, hateful, normal, spam &No \\
  \hline
  \citet{warner2012detecting} &Yahoo! &9k &No &anti-semitic, anti-black, anti-asian, anti-woman, anti-muslim, anti-immigrant, other-hate &No \\
  \hline
  \citet{nobata2016abusive} & Yahoo! &2k &No &clean, hate, derogatory, profanity &No \\
  \hline
  \citet{van2015detection} &Ask.fm &85k &No &threat/blackmail, insult, defamation, sexual talk, curse/exclusion, defense,  encouragement to the harasser &No \\
  \hline
  Ours &Reddit &22k  &Yes &hate, non-hate &Yes \\
  \hline
  Ours &Gab &34k  &Yes &hate, non-hate &Yes \\
  \hline
  
\end{tabular}
\caption{Comparison of our datasets with previous hate speech datasets. Conv.: Conversation. Interv.: Intervention.}
\label{tab:relatedwork}
\end{table*}

\section{Related Work}
\label{sec:related work}
In recent years, a few datasets for hate speech detection have been built and released by researchers. Most are collected from Twitter and are labeled using a combination of expert and non-expert hand labeling, or through machine learning assistance using a list of common negative words. It is widely accepted that labels can vary in their accuracy overall, though this can be mitigated by relying on a consensus rule to rectify disagreements in labels. A synopsis of these datasets can be found in Table~\ref{tab:relatedwork}.

~\citet{waseem2016hateful} collect 17k tweets based on hate-related slurs and users. The tweets are manually annotated with three categories: sexist (20.0\%), racist (11.7\%), and normal (68.3\%). 
Because the authors identified a number of prolific users during the initial manual search, the resulting dataset has a small number of users (1,236 users) involved, causing a potential selection bias. This problem is most prevalent on the 1,972 racist tweets, which are sent by only 9 Twitter users. To avoid this problem, we did not identify suspicious user accounts or utilize user information when collecting our data.

~\citet{davidson2017automated} use a similar strategy, which combines the utilization of hate keywords and suspicious user accounts to build a dataset from Twitter. But different from~\citet{waseem2016hateful}, this dataset consists of 25k tweets randomly sampled from the 85.4 million posts of a large number of users (33,458 users). 
This dataset is proposed mainly to distinguish hateful and offensive language, which tend to be conflated by many studies. 

~\citet{golbeck2017large} focus on online harassment on Twitter and propose a fine-grained labeled dataset with 6 categories. 
~\citet{founta2018large} introduce a large Twitter dataset with 100k tweets. 
Despite the large size of this dataset, the ratio of the hateful tweets are relatively low (5\%). Thus the size of the hateful tweets is around 5k in this dataset, which is not significantly larger than that of the previous datasets. 

The dataset introduced by~\citet{chatzakou2017mean} 
is different from the other datasets as it investigates the behavior of hate-related users on Twitter, instead of evaluating hate-related tweets. The large majority of the 1.5k users are labeled as spammers (31.8\%) or normal (60.3\%). Only a small fraction of the users are labeled as bullies (4.5\%) or aggressors (3.4\%). 

While most datasets are from single sources, ~\citet{george2017technology} introduce a dataset with a combination of Twitter (58.9\%), Reddit, and The Guardian. In total 20,432 unique comments were obtained with 4,136 labeled as harassment (20.2\%) and 16,296 as non-harassment (79.8\%).

Since most of the publicly available hate speech datasets are collected from Twitter, previous research of hate speech mainly focus on Twitter posts or users~\cite{waseem2016hateful,gao2017recognizing,burnap2016us,badjatiya2017deep,davidson2017automated}. While there are several studies on the other sources, such as Instagram~\cite{zhong2016content}, Yahoo!~\cite{warner2012detecting,nobata2016abusive}, and Ask.fm~\cite{van2015detection}, the hate speech on Reddit and Gab is not widely studied. What's more, all the previous hate speech datasets are built for the classification or detection of hate speech from a single post or user on social media, ignoring the context of the post and intervention methods needed to effectively calm down the users and diffuse negative online conversations.

\section{Dataset Collection}
\label{sec:dataset collection}
\subsection{Ethics}
\label{subsec:ethics}
Our study got approval from our Internal Review Board. Workers were warned about the offensive content before they read the data and they were informed by our instructions to feel free to quit the task at any time if they are uncomfortable with the content. Additionally, all personally identifiable information such as user names is masked in the datasets.

\subsection{Data Filtering}
\label{subsec:data filtering}
\begin{figure*}[t]
    \centering
    \includegraphics[width=1.0\linewidth]{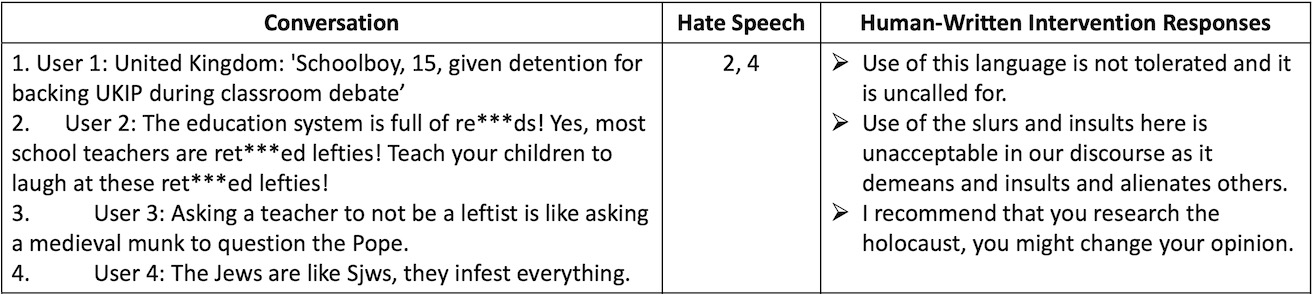}
    \caption{An example of the aggregated data. The first column is the conversation text. Indexes are added to each post. Indentations before each post indicate the structure of replies. The second column is the indexes of the human-labeled hateful post. Each bullet point in the third column is a human-written response.}
    \label{fig:conversation}
\end{figure*}

\textbf{Reddit:} To retrieve high-quality conversational data that would likely include hate speech, we referenced the list of the whiniest most low-key toxic subreddits\footnote{https://www.vice.com/en\_us/article/8xxymb/here-are-reddits-whiniest-most-low-key-toxic-subreddits}. Skipping the three subreddits that have been removed, we collect data from ten subreddits: \textit{r/DankMemes, r/Imgoingtohellforthis, r/KotakuInAction, r/MensRights, r/MetaCanada, r/MGTOW, r/PussyPass, r/PussyPassDenied, r/The\_Donald}, and \textit{r/TumblrInAction}. For each of these subreddits, we retrieve the top 200 hottest submissions using Reddit's API. To further focus on conversations with hate speech in each submission, we use hate keywords \cite{elsherief2018peer} to identify  potentially hateful comments and then reconstructed the conversational context of each comment. This context consists of all comments preceding and following a potentially hateful comment. 
Thus for each potentially hateful comment, we rebuild the conversation where the comment appears. Figure~\ref{fig:conversation} shows an example of the collected conversation, where the second comment contains a hate keyword and is considered as potentially hateful. Because a conversation may contain more than one comments with hate keywords, we removed any duplicated conversations.

\noindent\textbf{Gab:} We collect data from all the Gab posts in October 2018. Similar to Reddit, we use hate keywords~\cite{elsherief2018peer} to identify potentially hateful posts, rebuild the conversation context and clean duplicate conversations.

\subsection{Crowd-Sourcing}
\label{subsec:crowd-sourcing}
After we collected the conversations from both Reddit and Gab, we presented this data to Mechanical Turk workers to label and create intervention suggestions. In order not to over-burden the workers, we filtered out conversations consisting of more than 20 comments. Each assignment consists of 5 conversations. For Reddit, we also present the title and content of the corresponding submission in order to give workers more information about the topic and context. For each conversation, a worker is asked to answer two questions: 
\vspace{-8pt}
\begin{itemize}
    \item Q1: Which posts or comments in this conversation are hate speech?
    \item Q2: If there exists hate speech in the conversation, how would you respond to intervene? Write down a response that can probably hold it back (word limit: 140 characters).
\end{itemize} 
\vspace{-3pt}
If the worker thinks no hate speech exists in the conversation, then the answers to both questions are ``n/a''. To provide context, the definition of hate speech from Facebook\footnote{https://m.facebook.com/communitystandards/hate\_speech/}: ``We define hate speech as a direct attack on people based on what we call protected characteristics — race, ethnicity, national origin, religious affiliation, sexual orientation, caste, sex, gender, gender identity, and serious disease or disability.'' is presented to the workers. Also, to prevent workers from using hate speech in the response or writing responses that are too general, such as ``Please do not say that'', we provide additional instructions and rejected examples.

\subsection{Data Quality}
\label{subsec:data quality}
Each conversation is assigned to three different workers. To ensure data quality, we restrict the workers to be in an English speaking country including Australia, Canada, Ireland, New Zealand, the United Kingdom, and the United States, with a HIT approval rate higher than 95\%. Excluding the rejected answers, the collected data involves 926 different workers. The final hate speech labels (answers to Q1) are aggregated according to the majority of the workers' answers. A comment is considered hate speech only when at least two out of the three workers label it as hate speech. The responses (answers to Q2) are aggregated according to the aggregated result of Q1. If the worker's label to Q1 agrees with the aggregated result, then their answer to Q2 is included as a candidate response to the corresponding conversation but is otherwise disregarded.  See Figure~\ref{fig:conversation} for an example of the aggregated data.

\begin{figure}[t]
\centering

\includegraphics[width=1.0\linewidth]{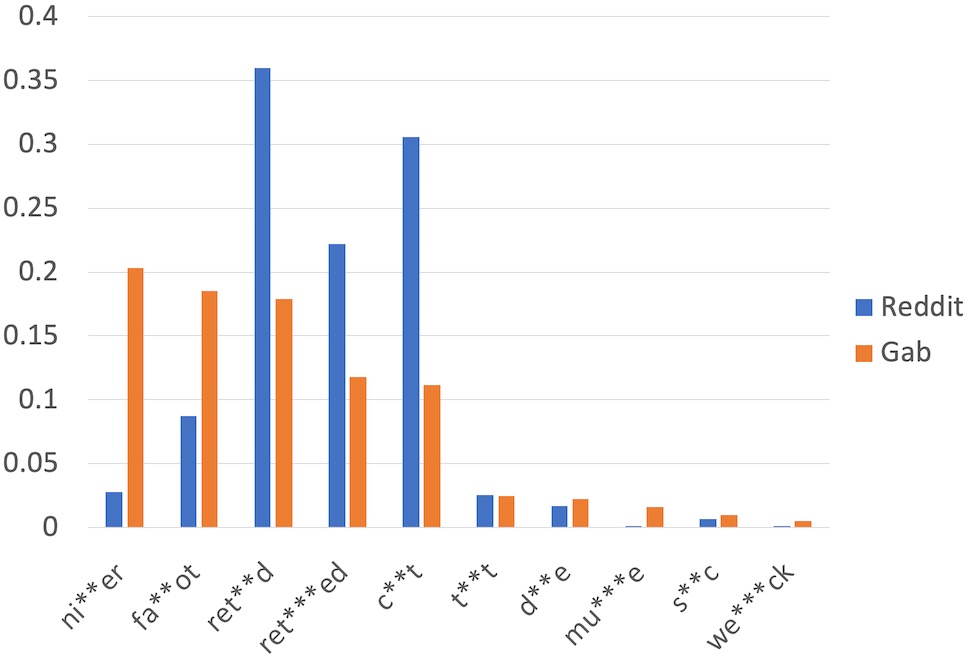}
\caption{The distributions of the top 10 keywords in the hate speech collected from Reddit and Gab. Hate keywords are masked.}
\label{fig:keyword distribution}
\end{figure}

\section{Dataset Analysis}
\label{sec:data analysis}
\subsection{Statistics}
\label{subsec:statistics}
From Reddit, we collected 5,020 conversations, including 22,324 comments. On average, each conversation consists of 4.45 comments and the length of each comment is 58.0 tokens. 5,257 of the comments are labeled as hate speech and 17,067 are labeled as non-hate speech. A majority of the conversations, 3,847 (76.6\%), contain hate speech. Each conversation with hate speech has 2.66 responses on average, for a total of 10,243 intervention responses. The average length of the intervention responses is 17.96 tokens. 

From Gab, we collected 11,825 conversations, consisting of 33,776 posts. On average, each conversation consists of 2.86 posts and the average length of each post is 35.6 tokens. 14,614 of the posts are labeled as hate speech and 19,162 are labeled as non-hate speech. Nearly all the conversations, 11,169 (94.5\%), contain hate speech. 31,487 intervention responses were originally collected for conversations with hate speech, or 2.82 responses per conversation on average. The average length of the intervention responses is 17.27 tokens.

Compared with the Gab dataset, there are fewer conversations and comments in the Reddit dataset, comments and conversations are longer, and the distribution of hate and non-hate speech labels is more imbalanced. Figure~\ref{fig:keyword distribution} illustrates the distributions of the top 10 keywords in the hate speech collected from Reddit and Gab separately. The Gab dataset and the Reddit dataset have similar popular hate keywords, but the distributions are very different. All the statistics shown above indicate that the characteristics of the data collected from these two sources are very different, thus the challenges of doing detection or generative intervention tasks on the dataset from these sources will also be different. 
\subsection{Intervention Strategies}
Removing duplicates, there are 21,747 unique intervention responses in the aggregated Gab dataset and 7,641 in the aggregated Reddit dataset. Despite the large diversity of the collected responses for intervention, we find workers tend to have certain strategies for intervention.

\vspace{1mm} \noindent\textbf{Identify Hate Keywords}: One of the most common strategies is to identify the inappropriate terms in the post and then urge the user to stop using that work.  For example, \textit{``The C word and language attacking gender is unacceptable. Please refrain from future use.''} This strategy is often used when the hatred in the post is mainly conveyed by specific hate keywords.

\vspace{1mm} \noindent\textbf{Categorize Hate Speech:} This is another common strategy used by the workers. The workers classify hate speech into different categories, such as racist, sexist, homophobic, etc. This strategy is often combined with identifying hate keywords or targets of hatred. For example, \textit{``The term ""fa**ot"" comprises homophobic hate, and as such is not permitted here.''}

\vspace{1mm} \noindent\textbf{Positive Tone Followed by Transitions:} This is a strategy where the response consists of two parts combined with a transitional word, such as ``but'' and ``even though''. The first part starts with affirmative terms, such as ``I understand'', ``You have the right to'', and ``You are free to express'', showing kindness and understanding, while the second part is to alert the users that their post is inappropriate. For example, \textit{``I understand your frustration, but the term you have used is offensive towards the disabled community. Please be more aware of your words.''}. Intuitively, compared with the response that directly warns, this strategy is likely more acceptable for the users and be more likely to clam down a quarrel full of hate speech.  

\vspace{1mm} \noindent\textbf{Suggest Proper Actions:} Besides warning and discouraging the users from continuing hate speech, workers also suggest the actions that the user should take. This strategy can either be combined with other strategies mentioned above or be used alone. In the latter case, a negative tone can be greatly alleviated. For example, \textit{``I think that you should do more research on how resources are allocated in this country.''}

\section{Generative Intervention}
\label{sec:methods}
Our datasets can be used for various hate speech tasks. In this paper, we focus on  generative hate speech intervention.

The goal of this task is to generate a response to hate speech that can mitigate its use during a conversation. The objective can be formulated as the following equation:
 \begin{equation}
    Obj=\max{}\sum_{(c,r)\in D}\log p(r|c)
 \end{equation}
 where $c$ is the conversation, $r$ is the corresponding intervention response, and $D$ is the dataset. 
 This task is closely related to the response generation and dialog generation, though several differences exist including dialog length, language cadence, and word imbalances. As a baseline, we chose the most common methods of these two tasks, such as Seq2Seq and VAE, to determine the initial feasibility of automatically generate intervention responses. More recent Reinforcement Learning method for dialog generation ~\cite{li2016deep} can also be applied to this task with slight modification. Future work will explore more complex, and unique models.

Similar to~\cite{li2016deep}, a generative model is considered as an agent. However, different from dialog generation, generative intervention does not have multiple turns of utterance, so the action of the agent is to select a token in the response. The state of the agent is given by the input posts and the previously generated tokens. Another result due to this difference is that the rewards with regard to ease of answering or information flow do not apply to this case, but the reward for semantic coherence does. Therefore, the reward of the agent is:
\begin{equation}
    rw(c,r)=\lambda_1\log p(r|c)+\lambda_2\log p_{back}(c|r)
\end{equation}
where $rw(c,r)$ is the reward with regard to the conversation $c$ and its reference response $r$ in the dataset. $p(r|c)$ denotes the probability of generating response $r$ given the conversation $c$, and $p_{back}(c|r)$ denotes the backward probability of generating the conversation based on the response, which is parameterized by another generation network. The reward is a weighted combination of these two parts, which are observed after the agent finishing generating the response. We refer the readers to~\citet{li2016deep} for details.

\section{Experiments}
\label{sec:experimens}
We evaluate the commonly-used detection and generation methods with our dataset. Due to the different characteristics of the data collected from the two sources (Section~\ref{sec:data analysis}), we treat them as two independent datasets.
\begin{figure*}[t]
\centering
\includegraphics[width=1.0\linewidth]{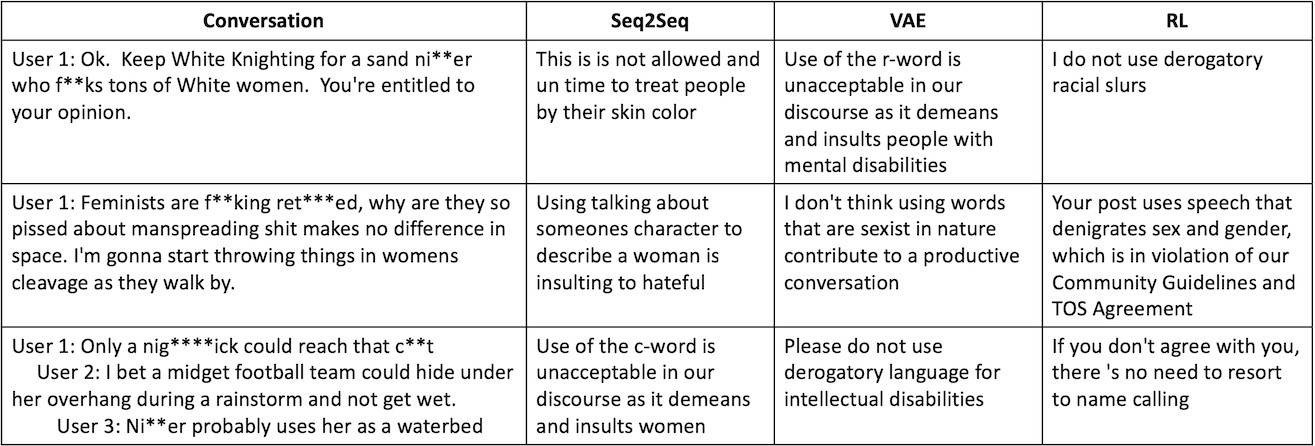}
\caption{Examples of the generated intervention responses. The hateful terms in the conversation are masked.}
\label{fig:generation samples}
\end{figure*}
\subsection{Experimental Settings}
\label{subsec:experimental settings}
For \textbf{binary hate speech detection}, we experimented the following four different methods. 
\vspace{1mm}

\noindent \textbf{Logistic Regression (LR):} We evaluate the Logistic  Regression  model  with  L2  regularization. The penalty  parameter  C  is  set  to  1. The input features  are  the  Term  Frequency  Inverse  Document Frequency (TF-IDF) values of up to 2-grams.  
\vspace{1mm}

\noindent\textbf{Support Vector Machine (SVM):} We evaluate the SVM model with linear kernels. We use L2 regularization and the coefficient is 1. The features are the same as in LR.
\vspace{1mm}

\noindent\textbf{Convolutional Neural Network (CNN):} We use the CNN model for sentence classification proposed by~\citet{kim2014convolutional} with default hyperparameters. The word embeddings are randomly initialized (CNN in Table~\ref{tab:detection results}) or initialized with pretrained Word2Vec~\cite{mikolov2013distributed} embeddings on Google News (CNN$^\ast$ in Table~\ref{tab:detection results}).
\vspace{1mm}

\noindent\textbf{Recurrent Neural Network (RNN):} The model we evaluated consists of 2-layer bidirectional Gated Recurrent Unit (GRU)~\cite{cho2014learning} followed by a linear layer. Same as for  CNN, we report the performance of RNN with two different settings of the word embeddings. 

The methods are evaluated on testing data randomly selected from the dataset with the ratio of 20\%. The input data  is not manipulated to manually balance the classes for any of the above methods. Therefore, the training and testing data retain the same distribution as the collected results (Section~\ref{sec:data analysis}). The methods are evaluated using F-1 score, Precision-Recall (PR) AUC, and Receiver-Operating-Characteristic (ROC) AUC.

For \textbf{generative hate speech intervention}, we evaluated the following three methods. 
\vspace{1mm}

\noindent\textbf{Seq2Seq}~\cite{sutskever2014sequence,cho2014learning}:  The encoder consists of 2 bidirectional GRU layers. The decoder consists of 2 GRU layers followed by a 3-layer MLP (Multi-Layer Perceptron).
\vspace{1mm}

\noindent\textbf{Variational Auto-Encoder (VAE)}~\cite{kingma2013auto}: The structure of the VAE model is similar to that of the Seq2Seq model, except that it has two independent linear layers followed by the encoder to calculate the mean and variance of the distribution of the latent variable separately. We assume the latent variable follows a multivariate Gaussian Distribution. KL annealing~\cite{klannealing} is applied during training.
\vspace{1mm}

\noindent\textbf{Reinforcement Learning (RL):} We also implement the Reinforcement Learning method described in Section~\ref{sec:methods}. The backbone of this model is the Seq2Seq model, which follows the same Seq2Seq network structure described above. This network is used to parameterize the probability of a response given the conversation. Besides this backbone Seq2Seq model, another Seq2Seq model is used to generate the backward probability. This network is trained in a similar way as the backbone Seq2Seq model, but with a response as input and the corresponding conversation as the target. In our implementation, the function of the first part of the reward ($\log p(r|c)$) is conveyed by the MLE loss. A curriculum learning strategy is adopted for the reward of $\log p_{back}(c|r)$ as in~\citet{ranzato2015sequence}. Same as in~\citet{li2016deep} and~\citet{ranzato2015sequence}, a baseline strategy is employed to estimate the average reward. We parameterize it as a 3-layer MLP. 

The Seq2Seq model and VAE model are evaluated under two different settings. In one setting, the input for the generative model is the complete conversation, while in the other setting, the input is the filtered conversation, which only includes the posts labeled as hate speech. The filtered conversation was necessary to test the Reinforcement Learning model, as it is too challenging for the backward model to reconstruct the complete conversation based only on the intervention response. 

In our experiments on the generative hate speech intervention task, we do not consider conversations without hate speech. 
The testing dataset is then randomly selected from the resulting dataset with the ratio of 20\%. Since each conversation can have multiple reference responses, we dis-aggregate the responses and construct a pair (conversation, reference response) for each of the corresponding references during training. Teacher forcing is used for each of the three methods. The automatic evaluation metrics include BLEU~\cite{papineni2002bleu}, ROUGE-L~\cite{lin2004rouge}, and METEOR~\cite{banerjee2005meteor}. 

\begin{table}[t!]
\centering
\small
\begin{tabular}{|l|c|c|c|c|c|c|}
  \hline
  Dataset & \multicolumn{3}{|c|}{Gab} &\multicolumn{3}{|c|}{Reddit}\\
  \hline
  Metric & F1 & PR & ROC &F1 &PR &ROC \\
  \hline
  LR &88.2 &94.5 &95.4 &64.7  &80.4  &91.4  \\
  \hline
  SVM &{88.6} &{94.7} &{95.6} &{75.7} &\textbf{81.1} &\textbf{92.0} \\
  \hline
  CNN  &87.5 &92.8 &92.6 &74.8 &76.8 &87.5 \\
  \hline
  RNN &87.6 &93.9 &94.2 &71.7 &76.1 &88.6 \\
  \hline
  CNN$^\ast$ &\textbf{89.6} &\textbf{95.2} & \textbf{95.8} & 76.9 & 80.1 & 90.9\\
  \hline
  RNN$^\ast$ &89.3 & 94.8 & 95.5 &\textbf{77.5} & 79.4 &90.6 \\
  \hline
\end{tabular}
\caption{Experimental results for the detection task. PR is Precision-Recall AUC and ROC is ROC AUC. The models marked with $^\ast$ use pretrained Word2Vec embeddings. The best results are in bold. }
\label{tab:detection results}
\end{table}
In order to validate and compare the quality of the generated results from each model, we also conducted human evaluations as previous research has shown that automatic evaluation metrics often do not correlate with human preference~\cite{socher2017summ}. We randomly sampled 450 conversations from the testing dataset. We then generated responses using each of the above models trained with the filtered conversation setting. In each assignment, a Mechanical Turk worker is presented 10 conversations, along with corresponding responses generated by the three models. For each conversation, the worker is asked to evaluate the effectiveness of the generated intervention by selecting a response that can best mitigate hate speech. 9 of the 10 questions are filled with the sampled testing data and the generated results, while the other is artificially constructed to monitor response quality. After selecting the 10 best mitigation measures, the worker is asked to select which of the three methods has the best diversity of responses over all the 10 conversations. Ties are permitted for answers. 
Assignments failed on the quality check are rejected.  
\subsection{Experimental Results and Discussion}
\label{subsec:detection results}
The experimental results of the detection task and the generative intervention task are shown in Table~\ref{tab:detection results} and Table~\ref{tab:generation results} separately. The results of the human evaluation are shown in Table~\ref{tab:human evalutation}. Figure~\ref{fig:generation samples} shows examples of the generated responses.

\begin{table*}[t!]
\centering
\small
\begin{tabular}{|l|c|c|c|c|c|c|c|c|c|c|c|c|}
  \hline
  Dataset & \multicolumn{6}{|c|}{Gab} &\multicolumn{6}{|c|}{Reddit}\\
  \hline
  Inp. Set.& \multicolumn{3}{|c|}{Complete}& \multicolumn{3}{|c|}{Filtered}& \multicolumn{3}{|c|}{Complete}& \multicolumn{3}{|c|}{Filtered}\\
  \hline
  Metric & B & R &M &B &R &M & B & R &M &B &R &M\\
  \hline
  Seq2Seq &\textbf{13.2}&\textbf{33.8}&{23.0} & \textbf{15.0} & \textbf{34.2} & 23.6 &{5.5} &\textbf{29.5} &{19.5}& 5.9  & 28.2  &20.0  \\
  \hline
  VAE &{12.2} &{32.5} &\textbf{23.4} & 12.4 & {32.8} &{21.8}  &\textbf{6.8} &{29.0} &\textbf{20.2} &\textbf{7.0} & \textbf{29.1} & \textbf{20.1}\\
  \hline
  RL  &-&-&- &{14.5} &33.1 &\textbf{23.9} &- &-&-&{4.4}&\textbf{29.1} &18.7 \\
  \hline
\end{tabular}
\caption{Experimental results for generative intervention task. Inp. Set.: Input Setting (Section~\ref{subsec:experimental settings}). B: BLEU. R: ROUGE-L. M: METEOR. Best results are in bold. }
\label{tab:generation results}
\end{table*}

\begin{table}[t!]
\centering
\small
\begin{tabular}{|l|c|c|c|c|}
  \hline
  Dataset & \multicolumn{2}{|c|}{Gab} &\multicolumn{2}{|c|}{Reddit}\\
  \hline
  Metric & Eff. &Div. &Eff. &Div.  \\
  \hline
  Seq2Seq Wins &22.4 &28.0 &\textbf{31.1} &\textbf{34.0}   \\
  \hline
  VAE Wins &20.0 &6.0  &26.0 &4.0 \\
  \hline
  RL Wins &\textbf{41.6} &\textbf{40.0} &30.0 &30.0 \\
  \hline
  Tie  &16.0 &26.0 &12.9 &32.0 \\
  \hline
\end{tabular}
\caption{Human evaluation results. Table values are the percentage of the answers. Eff.: Effectiveness, evaluates how well the generated responses can mitigate hate speech. Div: Diversity, evaluates how many different responses are generated. Best results are in bold.}
\label{tab:human evalutation}
\end{table}

As shown in Table~\ref{tab:detection results} and~\ref{tab:generation results}, all the classification and generative models perform better on the Gab dataset than on the Reddit dataset. We think this stems from the datasets' characteristics. First, the Gab dataset is larger and has a more balanced category distribution than the Reddit dataset. Therefore, it is inherently more challenging to train a classifier on the Reddit dataset. Further, the average lengths of the Reddit posts and conversations are much larger than those of Gab, potentially making the Reddit input nosier than the Gab input for both tasks. On both the Gab and Reddit datasets, the SVM classifier and the LR classifier achieved better performance than the CNN and RNN model with randomly initialized word embeddings. A possible reason is that without pretrained word embeddings, the neural network models tend to overfit on the dataset.

 For the generative intervention task, three models perform similarly on all three automatic evaluation metrics. As expected, the Seq2Seq model achieves higher scores with filtered conversation as input. However, this is not the case for the VAE model. This indicates that the two models may have different capabilities to capture important information in conversations. 

As shown in Table~\ref{tab:generation results}, applying Reinforcement Learning does not lead to higher scores on the three automatic metrics. However, human evaluation (Table~\ref{tab:human evalutation}) shows that the RL model creates responses that are potentially better at mitigating hate speech and are more diverse, which is consistent with~\citet{li2016deep}. There is a larger performance difference with the Gab dataset, while the effectiveness and the diversity of the responses generated by the Seq2Seq model and the RL model are quite similar on the Reddit dataset. One possible reason is that the size of the training data from Reddit (around 8k) is only 30\% the size of the training data from Gab.  
The inconsistency between the human evaluation results and the automatic ones indicates the automatic evaluation metrics listed in Table~\ref{tab:generation results} can hardly reflect the quality of the generated responses. As mentioned in Section~\ref{sec:data analysis}, annotators tend to have strategies for intervention. Therefore, generating the common parts of the most popular strategies for all the testing input can lead to high scores of these automatic evaluation metrics. For example, generating \textit{``Please do not use derogatory language.''} for all the testing Gab data can achieve 4.2 on BLEU, 20.4 on ROUGE, and 18.2 on METEOR. However, this response is not considered as high-quality because it is almost a universal response to all the hate speech, regardless of the context and topic.

Surprisingly, the responses generated by the VAE model have much worse diversity than the other two methods according to human evaluation. As indicated in Figure~\ref{fig:generation samples}, the responses generated by VAE tend to repeat the responses related to some popular hate keyword. For example, \textit{``Use of the r-word is unacceptable in our discourse as it demeans and insults people with mental disabilities.''} and \textit{``Please do not use derogatory language for intellectual disabilities.''} are the generated responses for a large part of the Gab testing data. According to Figure~\ref{fig:keyword distribution}, insults towards disabilities are the largest portion in the dataset, so we suspect that the performance of the VAE model is affected by the imbalanced keyword distribution.  

The sampled results in Figure~\ref{fig:generation samples} show that the Seq2Seq and the RL model can generate reasonable responses for intervention. However, as is to be expected with machine-generated text, in the other human evaluation we conducted, where Mechanical Turk workers were also presented with sampled human-written responses alongside the machine generated responses, the human-written responses were chosen as the most effective and diverse option a majority of the time (70\% or more) for both datasets. This indicates that there is significant room for improvement while generating automated intervention responses.

In our experiments, we only utilized the text of the posts, but more information is available and can be utilized, such as the user information and the title of a Reddit submission. 

\section{Conclusion}
\label{sec:conclusion}
Towards the end goal of mitigating the problem of online hate speech, we propose the task of generative hate speech intervention and introduce two fully-labeled datasets collected from Reddit and Gab, with crowd-sourced intervention responses. The performance of the three generative models: Seq2Seq, VAE, and RL, suggests ample opportunity for improvement. We intend to make our dataset freely available to facilitate further exploration of hate speech intervention and better models for generative intervention.

\section*{Acknowledgments}
This research was supported by the Intel AI Faculty Research Grant. The authors are solely responsible for the contents of the paper and the opinions expressed in this publication do not reflect those of the funding agencies.
\bibliography{emnlp-ijcnlp-2019}

\begin{thebibliography}{33}
\expandafter\ifx\csname natexlab\endcsname\relax\def\natexlab#1{#1}\fi

\bibitem[{Badjatiya et~al.(2017)Badjatiya, Gupta, Gupta, and
  Varma}]{badjatiya2017deep}
Pinkesh Badjatiya, Shashank Gupta, Manish Gupta, and Vasudeva Varma. 2017.
\newblock Deep learning for hate speech detection in tweets.
\newblock In \emph{Proceedings of the 26th International Conference on World
  Wide Web Companion}, pages 759--760. International World Wide Web Conferences
  Steering Committee.

\bibitem[{Banerjee and Lavie(2005)}]{banerjee2005meteor}
Satanjeev Banerjee and Alon Lavie. 2005.
\newblock Meteor: An automatic metric for mt evaluation with improved
  correlation with human judgments.
\newblock In \emph{Proceedings of the acl workshop on intrinsic and extrinsic
  evaluation measures for machine translation and/or summarization}, pages
  65--72.

\bibitem[{Bielefeldt et~al.(2011)Bielefeldt, La~Rue, and Muigai}]{ohchr2011}
Heiner Bielefeldt, Frank La~Rue, and Githu Muigai. 2011.
\newblock Ohchr expert workshops on the prohibition of incitement to national,
  racial or religious hatred.
\newblock \emph{Expert workshop on the Americas}.

\bibitem[{Bowman et~al.(2016)Bowman, Vilnis, Vinyals, Dai, J{\'{o}}zefowicz,
  and Bengio}]{klannealing}
Samuel~R. Bowman, Luke Vilnis, Oriol Vinyals, Andrew~M. Dai, Rafal
  J{\'{o}}zefowicz, and Samy Bengio. 2016.
\newblock Generating sentences from a continuous space.
\newblock In \emph{Proceedings of the 20th {SIGNLL} Conference on Computational
  Natural Language Learning, CoNLL 2016, Berlin, Germany, August 11-12, 2016},
  pages 10--21.

\bibitem[{Burnap and Williams(2016)}]{burnap2016us}
Pete Burnap and Matthew~L Williams. 2016.
\newblock {Us and Them: Identifying Cyber Hate on Twitter across Multiple
  Protected Characteristics}.
\newblock \emph{EPJ Data Science}, 5(1):11.

\bibitem[{Center(2017)}]{pew2017online}
Pew~Research Center. 2017.
\newblock Online harassment 2017.

\bibitem[{Chatzakou et~al.(2017)Chatzakou, Kourtellis, Blackburn,
  De~Cristofaro, Stringhini, and Vakali}]{chatzakou2017mean}
Despoina Chatzakou, Nicolas Kourtellis, Jeremy Blackburn, Emiliano
  De~Cristofaro, Gianluca Stringhini, and Athena Vakali. 2017.
\newblock Mean birds: Detecting aggression and bullying on twitter.
\newblock In \emph{Proceedings of the 2017 ACM on web science conference},
  pages 13--22. ACM.

\bibitem[{Cho et~al.(2014)Cho, van Merrienboer, G{\"{u}}l{\c{c}}ehre, Bahdanau,
  Bougares, Schwenk, and Bengio}]{cho2014learning}
Kyunghyun Cho, Bart van Merrienboer, {\c{C}}aglar G{\"{u}}l{\c{c}}ehre, Dzmitry
  Bahdanau, Fethi Bougares, Holger Schwenk, and Yoshua Bengio. 2014.
\newblock Learning phrase representations using {RNN} encoder-decoder for
  statistical machine translation.
\newblock In \emph{Proceedings of the 2014 Conference on Empirical Methods in
  Natural Language Processing, {EMNLP} 2014, October 25-29, 2014, Doha, Qatar,
  {A} meeting of SIGDAT, a Special Interest Group of the {ACL}}, pages
  1724--1734.

\bibitem[{Davidson et~al.(2017)Davidson, Warmsley, Macy, and
  Weber}]{davidson2017automated}
Thomas Davidson, Dana Warmsley, Michael Macy, and Ingmar Weber. 2017.
\newblock Automated hate speech detection and the problem of offensive
  language.
\newblock In \emph{Eleventh International AAAI Conference on Web and Social
  Media}.

\bibitem[{ElSherief et~al.(2018{\natexlab{a}})ElSherief, Kulkarni, Nguyen,
  Wang, and Belding}]{elsherief2018hate}
Mai ElSherief, Vivek Kulkarni, Dana Nguyen, William~Yang Wang, and Elizabeth
  Belding. 2018{\natexlab{a}}.
\newblock Hate lingo: A target-based linguistic analysis of hate speech in
  social media.
\newblock In \emph{Twelfth International AAAI Conference on Web and Social
  Media}.

\bibitem[{ElSherief et~al.(2018{\natexlab{b}})ElSherief, Nilizadeh, Nguyen,
  Vigna, and Belding}]{elsherief2018peer}
Mai ElSherief, Shirin Nilizadeh, Dana Nguyen, Giovanni Vigna, and Elizabeth
  Belding. 2018{\natexlab{b}}.
\newblock Peer to peer hate: Hate speech instigators and their targets.
\newblock In \emph{Twelfth International AAAI Conference on Web and Social
  Media}.

\bibitem[{Founta et~al.(2018)Founta, Djouvas, Chatzakou, Leontiadis, Blackburn,
  Stringhini, Vakali, Sirivianos, and Kourtellis}]{founta2018large}
Antigoni~Maria Founta, Constantinos Djouvas, Despoina Chatzakou, Ilias
  Leontiadis, Jeremy Blackburn, Gianluca Stringhini, Athena Vakali, Michael
  Sirivianos, and Nicolas Kourtellis. 2018.
\newblock Large scale crowdsourcing and characterization of twitter abusive
  behavior.
\newblock In \emph{Twelfth International AAAI Conference on Web and Social
  Media}.

\bibitem[{Gao et~al.(2017)Gao, Kuppersmith, and Huang}]{gao2017recognizing}
Lei Gao, Alexis Kuppersmith, and Ruihong Huang. 2017.
\newblock Recognizing explicit and implicit hate speech using a weakly
  supervised two-path bootstrapping approach.
\newblock In \emph{Proceedings of the Eighth International Joint Conference on
  Natural Language Processing (Volume 1: Long Papers)}, volume~1, pages
  774--782.

\bibitem[{Golbeck et~al.(2017)Golbeck, Ashktorab, Banjo, Berlinger, Bhagwan,
  Buntain, Cheakalos, Geller, Gergory, Gnanasekaran et~al.}]{golbeck2017large}
Jennifer Golbeck, Zahra Ashktorab, Rashad~O Banjo, Alexandra Berlinger,
  Siddharth Bhagwan, Cody Buntain, Paul Cheakalos, Alicia~A Geller, Quint
  Gergory, Rajesh~Kumar Gnanasekaran, et~al. 2017.
\newblock A large labeled corpus for online harassment research.
\newblock In \emph{Proceedings of the 2017 ACM on Web Science Conference},
  pages 229--233. ACM.

\bibitem[{Kennedy~III et~al.(2017)Kennedy~III, McCollough, Dixon, Bastidas,
  Ryan, Loo, and Sahay}]{george2017technology}
George~W. Kennedy~III, Andrew~W. McCollough, Edward Dixon, Alexie Bastidas,
  John Ryan, Chris Loo, and Saurav Sahay. 2017.
\newblock {Technology solutions to combat online harassment}.
\newblock In \emph{Proceedings of the first workshop on abusive language
  online}, pages 73--77.

\bibitem[{Kim(2014)}]{kim2014convolutional}
Yoon Kim. 2014.
\newblock Convolutional neural networks for sentence classification.
\newblock In \emph{Proceedings of the 2014 Conference on Empirical Methods in
  Natural Language Processing, {EMNLP} 2014, October 25-29, 2014, Doha, Qatar,
  {A} meeting of SIGDAT, a Special Interest Group of the {ACL}}, pages
  1746--1751.

\bibitem[{Kingma and Welling(2013)}]{kingma2013auto}
Diederik~P Kingma and Max Welling. 2013.
\newblock Auto-encoding variational bayes.
\newblock \emph{arXiv preprint arXiv:1312.6114}.

\bibitem[{Li et~al.(2016)Li, Monroe, Ritter, Jurafsky, Galley, and
  Gao}]{li2016deep}
Jiwei Li, Will Monroe, Alan Ritter, Dan Jurafsky, Michel Galley, and Jianfeng
  Gao. 2016.
\newblock Deep reinforcement learning for dialogue generation.
\newblock In \emph{Proceedings of the 2016 Conference on Empirical Methods in
  Natural Language Processing, {EMNLP} 2016, Austin, Texas, USA, November 1-4,
  2016}, pages 1192--1202.

\bibitem[{Lin(2004)}]{lin2004rouge}
Chin-Yew Lin. 2004.
\newblock Rouge: A package for automatic evaluation of summaries.
\newblock \emph{Text Summarization Branches Out}.

\bibitem[{Mikolov et~al.(2013)Mikolov, Sutskever, Chen, Corrado, and
  Dean}]{mikolov2013distributed}
Tomas Mikolov, Ilya Sutskever, Kai Chen, Greg~S Corrado, and Jeff Dean. 2013.
\newblock Distributed representations of words and phrases and their
  compositionality.
\newblock In \emph{Advances in neural information processing systems}, pages
  3111--3119.

\bibitem[{Nobata et~al.(2016)Nobata, Tetreault, Thomas, Mehdad, and
  Chang}]{nobata2016abusive}
Chikashi Nobata, Joel Tetreault, Achint Thomas, Yashar Mehdad, and Yi~Chang.
  2016.
\newblock Abusive language detection in online user content.
\newblock In \emph{Proceedings of the 25th international conference on world
  wide web}, pages 145--153. International World Wide Web Conferences Steering
  Committee.

\bibitem[{Ohlheiser(2016)}]{ohlheiser2016}
Abby Ohlheiser. 2016.
\newblock Banned from twitter? this site promises you can say whatever you
  want.
\newblock \emph{The Washington Post}.

\bibitem[{Papineni et~al.(2002)Papineni, Roukos, Ward, and
  Zhu}]{papineni2002bleu}
Kishore Papineni, Salim Roukos, Todd Ward, and Wei-Jing Zhu. 2002.
\newblock Bleu: a method for automatic evaluation of machine translation.
\newblock In \emph{Proceedings of the 40th annual meeting on association for
  computational linguistics}, pages 311--318. Association for Computational
  Linguistics.

\bibitem[{Paulus et~al.(2018)Paulus, Xiong, and Socher}]{socher2017summ}
Romain Paulus, Caiming Xiong, and Richard Socher. 2018.
\newblock A deep reinforced model for abstractive summarization.
\newblock In \emph{6th International Conference on Learning Representations,
  {ICLR} 2018}.

\bibitem[{Qian et~al.(2018{\natexlab{a}})Qian, ElSherief, Belding, and
  Wang}]{qian2018hierarchical}
Jing Qian, Mai ElSherief, Elizabeth Belding, and William~Yang Wang.
  2018{\natexlab{a}}.
\newblock Hierarchical cvae for fine-grained hate speech classification.
\newblock In \emph{Proceedings of the 2018 Conference on Empirical Methods in
  Natural Language Processing}, pages 3550--3559.

\bibitem[{Qian et~al.(2018{\natexlab{b}})Qian, ElSherief, Belding, and
  Wang}]{qian2018leveraging}
Jing Qian, Mai ElSherief, Elizabeth Belding, and William~Yang Wang.
  2018{\natexlab{b}}.
\newblock Leveraging intra-user and inter-user representation learning for
  automated hate speech detection.
\newblock In \emph{Proceedings of the 2018 Conference of the North American
  Chapter of the Association for Computational Linguistics: Human Language
  Technologies, Volume 2 (Short Papers)}, volume~2, pages 118--123.

\bibitem[{Ranzato et~al.(2016)Ranzato, Chopra, Auli, and
  Zaremba}]{ranzato2015sequence}
Marc'Aurelio Ranzato, Sumit Chopra, Michael Auli, and Wojciech Zaremba. 2016.
\newblock Sequence level training with recurrent neural networks.
\newblock In \emph{4th International Conference on Learning Representations,
  {ICLR} 2016}.

\bibitem[{Schmidt and Wiegand(2017)}]{shmidt2017survey}
Anna Schmidt and Michael Wiegand. 2017.
\newblock {A survey on hate speech detection using natural language
  processing}.
\newblock In \emph{Proceedings of the Fifth International Workshop on Natural
  Language Processing for Social Media}, pages 1--10.

\bibitem[{Sutskever et~al.(2014)Sutskever, Vinyals, and
  Le}]{sutskever2014sequence}
Ilya Sutskever, Oriol Vinyals, and Quoc~V Le. 2014.
\newblock Sequence to sequence learning with neural networks.
\newblock In \emph{Advances in neural information processing systems}.

\bibitem[{Van~Hee et~al.(2015)Van~Hee, Lefever, Verhoeven, Mennes, Desmet,
  De~Pauw, Daelemans, and Hoste}]{van2015detection}
Cynthia Van~Hee, Els Lefever, Ben Verhoeven, Julie Mennes, Bart Desmet, Guy
  De~Pauw, Walter Daelemans, and V{\'e}ronique Hoste. 2015.
\newblock {Detection and Fine-grained Classification of Cyberbullying Events}.
\newblock In \emph{RANLP'15: International Conference Recent Advances in
  Natural Language Processing}, pages 672--680.

\bibitem[{Warner and Hirschberg(2012)}]{warner2012detecting}
William Warner and Julia Hirschberg. 2012.
\newblock {Detecting Hate Speech on the World Wide Web}.
\newblock In \emph{ACL'12: Proceedings of the 2nd Workshop on Language in
  Social Media}, pages 19--26. Association for Computational Linguistics.

\bibitem[{Waseem and Hovy(2016)}]{waseem2016hateful}
Zeerak Waseem and Dirk Hovy. 2016.
\newblock Hateful symbols or hateful people? predictive features for hate
  speech detection on twitter.
\newblock In \emph{Proceedings of the NAACL student research workshop}, pages
  88--93.

\bibitem[{Zhong et~al.(2016)Zhong, Li, Squicciarini, Rajtmajer, Griffin,
  Miller, and Caragea}]{zhong2016content}
Haoti Zhong, Hao Li, Anna~Cinzia Squicciarini, Sarah~Michele Rajtmajer,
  Christopher Griffin, David~J Miller, and Cornelia Caragea. 2016.
\newblock {Content-Driven Detection of Cyberbullying on the Instagram Social
  Network.}
\newblock In \emph{IJCAI'16: Proceedings of the 25th International Joint
  Conference on Artificial Intelligence}, pages 3952--3958.

\end{thebibliography}
\bibliographystyle{acl_natbib}
\appendix

\end{document}